\documentclass[11pt,a4paper]{article}
\usepackage[nohyperref]{acl2019}
\usepackage{times}
\usepackage{latexsym}
\usepackage{algpseudocode}
\usepackage{algorithm}
\usepackage{booktabs}
\usepackage{graphicx}
\usepackage{setspace}

\usepackage{url}

\aclfinalcopy % Uncomment this line for the final submission
\def\aclpaperid{11} %  Enter the acl Paper ID here

\algtext*{EndFor}
\algtext*{EndFunction}
\algtext*{EndIf}
\algrenewcommand\algorithmicindent{1.0em}
\title{Towards Unsupervised Grammatical Error Correction \\ using Statistical
Machine Translation with Synthetic Comparable Corpus}

\author{Satoru Katsumata \and Mamoru Komachi \\
  Tokyo Metropolitan University \\
    {\tt katsumata-satoru@ed.tmu.ac.jp},
    {\tt komachi@tmu.ac.jp}}

\date{}

\begin{document}
\maketitle

\begin{abstract}
  %This paper describes our grammatical error correction (GEC) system for the shared task at Building Educational Applications 2019 (BEA2019).
  %Most work in grammatical error correction relies on a large-scale training data.
  %However, large-scale learner corpora annotated with corrections are difficult to obtain in many languages.
  We introduce unsupervised techniques based on phrase-based statistical machine translation for grammatical error correction (GEC) trained on a pseudo learner corpus created by Google Translation.
  We verified our GEC system through experiments on various GEC dataset, including a low resource track of the shared task at Building Educational Applications 2019 (BEA2019).
  As a result, we achieved an $\mathrm{F_{0.5}}$ score of 28.31 points with the test data of the low resource track.
\end{abstract}

\section{Introduction}
Research on grammatical error correction (GEC) has gained considerable attention recently.
Many studies treat GEC as a task that involves translation from a grammatically erroneous sentence (source-side) into a correct sentence (target-side) and thus, leverage methods based on machine translation (MT) for GEC.
For instance, some GEC systems use large parallel corpora and synthetic data (\citeauthor{ge2018}, \citeyear{ge2018}; \citeauthor{xie2018}, \citeyear{xie2018}). 

%One such large-scale parallel corpus is the public parallel corpus extracted from Lang-8.
%, which is a social media for learning and practicing foreign languages that involves mutual corrections being made for second language learners by native language speakers.   
%This corpus consists of sentences written by language learners on the source side and sentences corrected by native speakers on the target side.
%However, the sentence on the target side is not necessarily completely corrected because annotators are learners of other languages; hence, they are not always experts in teaching.
%It is costly to have experts correct erroneous sentences.

We introduce an unsupervised method based on MT for GEC that does not almost use parallel learner data.
In particular, we use methods proposed by \citet{marie-fujita}, \citet{artetxe_usmt}, and \citet{lample}.
These methods are based on phrase-based statistical machine translation (SMT) and two phrase table refinements, i.e., forward and backward refinement.
Forward refinement simply arguments a learner corpus with automatic corrections whereas backward refinement expends both source-side and target-side data to train GEC model using back-translation \citep{back-translation}. 
%If we can use monolingual data written by native speakers as the target side, we will be able to exploit an unannotated learner corpus for GEC. 
%Forward refinement used by \citet{marie-fujita} simply augments a learner corpus with automatic corrections.
%whereas backward refinement expends both a learner corpus and a raw corpus to train GEC model using back-translation \citep{back-translation}.
%We also use forward refinement for improvement of phrase table.

Unsupervised MT techniques do not require a parallel but a comparable corpus as training data.
Therefore, we use comparable translated texts using Google Translation as the source-side data. 
Specifically, we use News Crawl written in English as target-side data and News Crawl written in another language translated into English as source-side data.

We identified the difference between forward and backward refinement with CoNLL-2014 dataset and JFLEG dataset; 
 the former generates fluent outputs.
 % and the latter makes highly adequate corrections.
We also verified our GEC system through experiments for a low resource track of the shared task at Building Educational Applications 2019 (BEA2019). 
The experimental results show that our system achieved an $\mathrm{F_{0.5}}$ score of 28.31 points in the low resource track of the shared task at BEA2019.
%The main contributions of this study are as follows:
%\begin{enumerate}
%\item To the best of our knowledge, this research is the first attempt to investigate whether a learner corpus needs to be annotated with errors and corrections.
%\item We obtained an $\mathrm{F_{0.5}}$ score of 22.40 points with the CoNLL-2014 dataset and a GLEU score of 43.85 points with the JFLEG dataset for English GEC.
%\item We determined the difference in correction via forward and backward refinement; the former generates fluent outputs and the latter makes highly adequate corrections.
%\end{enumerate}
	
\section{Unsupervised GEC}
Algorithm \ref{algorithm} shows the pseudocode for unsupervised GEC.
This code is derived from \citet{artetxe_usmt}. 
First, the cross-lingual phrase embeddings are acquired.
Second, a phrase table is created based on these cross-lingual embeddings.
Third, the phrase table is combined with a language model trained by monolingual data to initialize a phrase-based SMT system.
Finally, the SMT system is updated through iterative forward-translation or back-translation.

\paragraph{Cross-lingual embeddings}
First, $n$-gram embeddings were created on the source- and target-sides.
Specifically, each monolingual embedding was created based on the source- and target-sides using a variant of skip-gram \citep{mikolov2013_skip} for unigrams, bigrams, and trigrams with high frequency\footnote{We used the most frequent 200K unigrams, 400K bigrams, and 400K trigrams in the monolingual data.} in the monolingual data.
Next, the monolingual embeddings were mapped onto a shared space to obtain cross-lingual embeddings.
The self-learning method of \citet{artetxe_cross} was used for unsupervised mapping.

\begin{algorithm}[t]
  \caption{Unsupervised GEC}
  \label{algorithm}
  \small
  \begin{spacing}{1.12}
  \begin{algorithmic}[1]
    \Require language model of the source-side $LM_s$
    \Require language models of the target-side $LM_t$
    \Require source training corpus  $C_s$
    \Require target training corpus  $C_t$
    \Require tuning data $T$
    \Require iteration number $N$
    \Ensure source-to-target phrase table $P^{(N)}_{s \to t}$
    \Ensure source-to-target weights $W^{(N)}_{s \to t}$
    \State $W_s^{emb} \gets \Call{train}{C_s}$
    \State $W_t^{emb} \gets \Call{train}{C_t}$
    \State $W_s^{cross\_emb}, W_t^{cross\_emb} \gets \Call{mapping}{W_s^{emb}, W_t^{emb}}$
    \If{forward\_refinement}
      \State $P^{(0)}_{s \to t}$ $\gets$ \Call{initialize}{$W_s^{cross\_emb}, W_t^{cross\_emb}$}
      \State $W^{(0)}_{s \to t}$ $\gets$ \Call{tune}{$P^{(0)}_{s \to t}, LM_t, T$}
      \For {$iter = 1, \dots, N$} \State $\mathrm{synthetic\_data}_t$ 
        \State \hspace{\algorithmicindent} $\gets$ \Call{decode}{$P^{(iter-1)}_{s \to t}$, $LM_t$, $W^{(iter-1)}_{s \to t}$, $C_s$}
        \State $P^{(iter)}_{s \to t}$ $\gets$ \Call{train}{$C_s$, $\mathrm{synthetic\_data}_t$}
        \State $W^{(iter)}_{s \to t} \gets \Call{tune}{P^{(iter)}_{s \to t}, LM_t, T}$ 
      \EndFor
    \ElsIf{backward\_refinement}
      \State $P^{(0)}_{t \to s}$ $\gets$ \Call{initialize}{$W_t^{cross\_emb}, W_s^{cross\_emb}$}
      \State $W^{(0)}_{t \to s}$ $\gets$ \Call{tune}{$P^{(0)}_{t \to s}, LM_s, T$}
      \For {$iter = 1, \dots, N$} 
        %\State \mbox{$\mathrm{synthetic\_data}_s$ $\gets$ \Call{decode}{$P^{(iter-1)}_{t \to s}$, $LM_s$, $C_t$}}
        \State $\mathrm{synthetic\_data}_s$
        \State \hspace{\algorithmicindent} $\gets$ \Call{decode}{$P^{(iter-1)}_{t \to s}$, $LM_s$, $W^{(iter-1)}_{t \to s}$, $C_t$}
        \State $P^{(iter)}_{s \to t}$ $\gets$ \Call{train}{$\mathrm{synthetic\_data}_s$, $C_t$}
        \State $W^{(iter)}_{s \to t} \gets \Call{tune}{P^{(iter)}_{s \to t}, LM_t, T}$ 
        \State $\mathrm{synthetic\_data}_t$ 
        \State \hspace{\algorithmicindent} $\gets$ \Call{decode}{$P^{(iter-1)}_{s \to t}$, $LM_t$, $W^{(iter-1)}_{s \to t}$, $C_s$}
        \State $P^{(iter)}_{t \to s}$ $\gets$ \Call{train}{$\mathrm{synthetic\_data}_t$, $C_s$}
        \State $W^{(iter)}_{t \to s} \gets \Call{tune}{P^{(iter)}_{t \to s}, LM_s, T}$ 
      \EndFor
    \EndIf
    %\State $P^{(N)}_{s \to t} \gets \Call{refine\_pt}{N, P^{(0)}_{s \to t}, P^{(0)}_{t \to s}, LM_s, LM_t, C_s, C_t}$
  \end{algorithmic}
  \end{spacing}
\end{algorithm}

\paragraph{Phrase table induction}
A phrase table was created based on the cross-lingual embeddings.
In particular, this involved the creation of phrase translation models and lexical translation models.

The translation candidates were limited in the source-to-target phrase translation model $\phi (\overline{f} | \overline{e})$ for each source phrase $\overline{e}$ to its 100 nearest neighbor phrases $\overline{f}$ on the target-side.
The score of the phrase translation model was calculated based on the normalized cosine similarity between the source and target phrases.
$$ \phi ( \overline { f } | \overline { e } ) = \frac { \exp (\cos ( \overline { e }, \overline { f } ) / \tau) } { \sum _ { \overline { f } ^ { \prime } } \exp (\cos ( \overline { e }, \overline { f } ^ { \prime } ) / \tau) } \eqno(1) $$
$\overline{f}^\prime$ represents each phrase embedding on the target-side and $\tau$ is a temperature parameter that controls the confidence of prediction\footnote{As in \citet{artetxe_usmt}, $\tau$ is estimated by maximizing the phrase translation probability between an embedding and the nearest embedding on the opposite side.}. 
The backward phrase translation probability $\phi (\overline{e} | \overline{f})$ was determined in a similar manner.

The source-to-target lexical translation model $\mathrm{lex}(\overline{f} | \overline{e})$ considers the word with the highest translation probability in a target phrase for each word in a source phrase.
The score of the lexical translation model was calculated based on the product of respective phrase translation probabilities.
$$
\mathrm { lex } ( \overline { f } | \overline { e } ) = \prod _ { i } \max \left( \epsilon, \max _ { j } \phi \left( \overline { f } _ { i } | \overline { e } _ { j } \right) \right)
\eqno(2)
$$
$\epsilon$ is a constant term for the case where no alignments are found.
As in \citet{artetxe_usmt}, the term was set to 0.001.
The backward lexical translation probability $\mathrm{lex}(\overline{e} | \overline{f})$ is calculated in a similar manner.

%\begin{algorithm}[t]
%  \caption{Refinement of Phrase Table}
%  \label{algorithm_refine}
%  \small
%  \begin{algorithmic}
%    \Function{forward-refine}{$N, P^{(0)}_{s \to t}, LM_t, C_s$}
%    \For {$iter = 1, \dots, N$}
%      \State \mbox{$\mathrm{synthetic\_data}_t$ $\gets$ \Call{decode}{$P^{(iter-1)}_{s \to t}$, $LM_t$, $C_s$}}
%      \State $P^{(iter)}_{s \to t}$ $\gets$ \Call{train}{$C_s$, $\mathrm{synthetic\_data}_t$}
%    \EndFor
%    \State \Return $P^{(N)}_{s \to t}$
%    \EndFunction
%    \Function{\mbox{backward-refine}}{$N, P^{(0)}_{t \to s}, LM_s, LM_t, C_s, C_t$}
%    \For {$iter = 1, \dots, N$} 
%      \State \mbox{$\mathrm{synthetic\_data}_s$ $\gets$ \Call{decode}{$P^{(iter-1)}_{t \to s}$, $LM_s$, $C_t$}}
%      \State $P^{(iter)}_{s \to t}$ $\gets$ \Call{train}{$\mathrm{synthetic\_data}_s$, $C_t$}
%      \State $\mathrm{synthetic\_data}_t$ $\gets$ \Call{decode}{$P^{(iter)}_{s \to t}$, $LM_t$, $C_s$}
%      \State $P^{(iter)}_{t \to s}$ $\gets$ \Call{train}{$\mathrm{synthetic\_data}_t$, $C_s$}
%    \EndFor
%    \State \Return $P^{(N)}_{s \to t}$
%    \EndFunction
%  \end{algorithmic}
%\end{algorithm}

\paragraph{Refinement of SMT system}
The phrase table created is considered to include noisy phrase pairs.
Therefore, we update the phrase table using an SMT system.
The SMT system trained on synthetic data eliminates the noisy phrase pairs using language models trained on the target-side corpus.
%Algorithm \ref{algorithm_refine} presents this process.
This process corresponds to lines 4---23 in Algorithm \ref{algorithm}.
%The phrase table is refined with forward refinement \citep{marie-fujita}.
The phrase table can be refined in either of two ways: forward and backward refinement.

% 何をしているかの説明
For forward refinement \citep{marie-fujita}, target synthetic data were generated from the source monolingual data using the source-to-target phrase table $P_{s \to t}^{(0)}$ and target language models $LM_t$.
A new phrase table $P_{s \to t}^{(1)}$ was then created with this target synthetic corpus.
This operation was executed $N$ times.

For backward refinement \citep{artetxe_usmt}, source synthetic data were generated from the target monolingual data using the target to source phrase table $P_{t \to s}^{(0)}$ and source language model $LM_s$.
A new source to target phrase table $P_{s \to t}^{(1)}$ was created with this source synthetic parallel corpus.
Next, target synthetic data were generated from the source monolingual data using $P_{s \to t}^{(1)}$ and target language model $LM_t$.
The target to source phrase table $P_{t \to s}^{(1)}$ was built using this target synthetic data.

\paragraph{Construction of a comparable corpus}
This unsupervised method is based on the assumption that the source and target corpora are comparable.
In fact, \citet{lample}, \citet{artetxe_usmt} and \citet{marie-fujita} use the News Crawl of source and target language as training data.
%Therefore, we make and use a comparable corpus.

To make a comparable corpus for GEC, we use translated texts using Google Translation as the source-side data.
Specifically, we use Finnish News Crawl translated into English as source-side.
English News Crawl is used as the target-side as is.
Finnish data is used because Finnish is not similar to English.

This translated data does not include misspelled words.
To address these words, we use a spell checker as a preprocessing step before inference.

\begin{table}[t]
  \centering
  \small
  \begin{tabular}{lrr} \toprule
    Corpus & Sent. & Learner \\ \midrule \midrule
    Training data & & \\ 
    \hspace{\algorithmicindent} Fi News Crawl & 1,904,880 & No \\
    \hspace{\algorithmicindent} En News Crawl & 2,116,249 & No\\
    \hspace{\algorithmicindent} One-Billion & 24,482,651 & No\\ \midrule
    for CoNLL-14 & & \\
    \hspace{\algorithmicindent} CoNLL-13 & 1,312 & Yes \\
    for JFLEG test & & \\
    \hspace{\algorithmicindent} JFLEG dev & 754 & Yes \\
    \multicolumn{2}{l}{for W\&I+LOCNESS (BEA2019)} & \\
    \hspace{\algorithmicindent} tuning data & 2,191 & Yes\\
    \hspace{\algorithmicindent} dev data & 2,193 & Yes \\ \bottomrule
    \end {tabular}
    \caption{Data statics: train and dev data size.}
  \label{data}
\end{table}

%\vspace{-1mm}
\section{Experiment of low resource GEC}
\subsection{Experimental setting}
Table \ref{data} shows the training and development data size.
Unless mentioned otherwise, Finnish News Crawl 2014---2015 translated into English was used as source training data and English News Crawl 2017 was used as target training data.
To train the extra language model of the target-side ($LM_t$), we used training data of One Billion Word Benchmark \citep{one-billion}.
We used \verb|googletrans v2.4.0|\footnote{https://github.com/ssut/py-googletrans} for Google Translation 
 and obtained 2,122,714 translated sentences.
% \footnote{Finnish News Crawl 2014---2015 has 6,360,479 sentences but the googletrans module sometimes did not work well.}.
We sampled the 3,000,000 sentences from English News Crawl 2017 and excluded the sentences with more than 150 words for either source- and target-side data.
Finally, the synthetic comparable corpus comprises processed News Crawl data listed in Table \ref{data}.

Our system was verified using three GEC datasets, CoNLL-14 \citep{ng2014}, JFLEG test set \citep{jfleg} and W\&I+LOCNESS (\citeauthor{bea2019}, \citeyear{bea2019}; \citeauthor{locness}, \citeyear{locness}).
We used the CoNLL-13 dataset \citep{conll13} and JFLEG dev set as tuning data for CoNLL-14 and JFLEG test, respectively.
The low resource track at BEA2019 permitted to use W\&I+LOCNESS development set, so we split it in half; tune data and dev data\footnote{Because W\&I+LOCNESS data has four types of learner level, we split it so that each learner level is equal.}.

These data were tokenized by \verb|spaCy v1.9.0|\footnote{https://github.com/explosion/spaCy} and the \verb|en_core_web_sm-1.2.0| model for W\&I+LOCNESS.
For CoNLL-14 and JFLEG test set, \verb|NLTK| \citep{nltk} tokenizer was used.
We used moses truecaser for the training data; this truecaser model was learned from processed English News Crawl.
We used byte-pair-encoding \citep{bpe} learned from processed English News Crawl; the number of operations was 50K.

The implementation made by \citet{artetxe_usmt}\footnote{https://github.com/artetxem/monoses}  was modified to conduct the experiments.
Specifically, some features were added; word-level Levenshtein distance, word-, and character-level edit operation, operation sequence model \citep{osm}\footnote{Operation sequence model was used in the refinement step.}, and 9-gram word class language model\footnote{
This class language model was not used when training target to source model.}, 
 similar to \citet{hybrid-gec} without sparse features.
Word class language model was trained with One Billion Word Benchmark data; the number of classes is 200, and the word class was estimated with \verb|fastText| \citep{fasttext}.
The distortion feature was not used.

\verb|Moses| \citep{moses} was used to train the SMT system.
\verb|FastAlign| \citep{dyer2013} was used for word alignment and \verb|KenLM| \citep{kenlm} was used to train the 5-gram language model over each processed English News Crawl and One Billion Word Benchmark.
%MERT \citep{mert} was used with the tuning data of CoNLL-13 and W\&I+LOCNESS for \verb|M^2 Scorer| \citep{max_match} and the tuning data of JFLEG for \verb|GLEU| \citep{gleu}.
MERT \citep{mert} was used with \verb|M^2 Scorer| \citep{max_match} for the tuning data of CoNLL-13 and W\&I+LOCNESS and with \verb|GLEU| \citep{gleu} for the tuning data of JFLEG.
Synthetic sentence pairs with a [3, 80] sentence length were used at the refinement step.
The number of iterations $N$ was set to 3 or 5, and the embedding dimension was set to 300.
For the low resource track, we decided best iteration of forward refinement with the dev data and submitted the output of the best iteration model.

We used \verb|pyspellchecker|\footnote{https://github.com/barrust/pyspellchecker} as a spell checker.
This tool uses Levenshtein distance to obtain permutations within an edit distance of 2 over the words included in a word list.
We made the word list from One Billion Word Benchmark and included words that occur more than five times.

For comparison, supervised GEC with SMT and neural MT (NMT) were conducted using the data extracted from Lang-8 \citep{mizumoto2011} as for CoNLL-14 and JFLEG.
In supervised SMT, the feature weights were tuned and the setting was the same as that in unsupervised SMT (USMT).
In supervised NMT, a convolutional Encoder--Decoder model \citep{convs2s} was used and the parameter settings were similar to those in \citet{ge2018}.

We report precision, recall, and $\mathrm{F_{0.5}}$ score for CoNLL-14 and W\&I+LOCNESS data and GLEU score for JFLEG test set.
% based on the dev data and official test data. citecitecite これ絶対どっかに入れて
The output of CoNLL-14 and W\&I+LOCNESS dev data was evaluated using $\mathrm{M^2}$ scorer and ERRANT scorer \citep{errant}, respectively.
%The output of dev data was evaluated using ERRANT scorer \citep{errant} similarly to official test data.

%\begin{table*}[t]
%   \centering
%   \small
%   \begin{tabular}{lp{14.4cm}} \toprule
   % 14.4or14.7
%   src & It is better to let the other party know the fact than after the baby is born and certain type of genetic disease is found. \\ \midrule \midrule
%   iter 0 & When you want to kill people know now than before the stroller and certain types of societal disease corrected. \\ \midrule
%   iter 1 & It is better to let the other party. The fact than after the baby is born, and \textbf{a} certain type of genetic cancer was found. \\ \midrule
%   iter 2, 3 & It is better to let the other party know the fact than after the baby is born and \textbf{a} certain type of genetic disease is found. \\\midrule
%   gold & It is better to let the other party know the \textbf{facts} \textbf{rather than wait until} after the baby is born and \textbf{a} certain type of genetic disease is found. \\\bottomrule
%   \end{tabular}
%      \caption{An example of USMT$_{\mathrm{backward}}$ in various iterations with CoNLL-14.}
 %  \label{result_example}
 %\end{table*}

\begin{table}[!t]
  \centering
  \scalebox{0.8}{
  %\small
  \begin{tabular}{lrrrrr} \toprule
      &   &  \multicolumn{3}{c}{CoNLL-14 ($\mathrm{M^2}$)} &JFLEG  \\ \cmidrule(lr){3-5} \cmidrule(lr){6-6}
      & iter & P & R & $\mathrm{F_{0.5}}$ & GLEU   \\ \midrule \midrule
    No edit & - & - & - & - & 40.54 \\ \midrule
    Supervised NMT & - & \underline{53.11} & 26.47 & \underline{44.21} & 54.04 \\ \midrule
    Supervised SMT & - & 43.02 & 33.18 & 40.61 & \underline{55.93}  \\ \midrule
    Unsupervised SMT & 0 & 21.82 & \underline{\textbf{36.75}} & 23.75 & 49.94 \\
    w/ forward\_refine & 1 & \textbf{25.92} & 32.65 & \textbf{27.04} & \textbf{50.65} \\
     & 2 & 25.58 & 31.02 & 26.51 & 50.19 \\
     & 3 & 23.95 & 33.13 & 24.54 & 50.40 \\ \midrule
    w/ backward\_refine & 1 & 22.39 & 33.39 & 23.97 & 49.02 \\
     & 2 & 24.96 & 27.13 & 25.36 & 48.90\\
     & 3 & 26.07 & 21.01 & 24.87 & 48.75 \\ \bottomrule
  \end{tabular}
  }
    \caption{$\mathrm{M^2}$ and GLEU results. The bold scores represent the best score in unsupervised SMT. The underlined scores represent the best overall score.}
  \label{result_score_conll}
\end{table}

\subsection{CoNLL-14 and JFLEG Results}\label{conll}
Table \ref{result_score_conll} shows the results of the GEC experiments for CoNLL-14 and JFLEG. 
The $\mathrm{F_{0.5}}$ score for USMT$_{\mathrm{forward}}$ in iter 1 is 13.57 points lower than that of supervised SMT and 17.17 points lower than that of supervised NMT.
On JFLEG, the highest score was achieved with USMT$_{\mathrm{forward}}$ in iter 1 among the unsupervised SMT models; its GLEU scores are 5.28 points and 3.39 points lower than those of supervised SMT and supervised NMT, respectively.

According to the improvement of iteration from 0 to 1, it is confirmed that the forward refinement works well.
However, it is observed that the system with forward refinement ceases to improve after iteration 1. 
In forward refinement, the source-side data is not changed, and target-side data is generated from the source-side for each iteration.
Therefore, the quality of the source-side data is important for this refinement method.
In this study, we use the automatically translated text as source-side data; thus, it is considered that the quality is not high and the refinement after iteration 1 does not work.

\begin{table}[t]
  \centering
  %\scalebox{0.8}{
  \begin{tabular}{lrrr} \toprule
    Src & Precision & Recall & $\mathrm{F_{0.5}}$ \\ \midrule
    Fi News Crawl & 29.17 & 28.52 & 29.04 \\
    Ru News Crawl & 27.11 & 29.84 & 27.62 \\
    Fr News Crawl & 25.05 & 30.27 & 25.94 \\
    De News Crawl & 23.26 & 26.04 & 25.04 \\ \bottomrule
  \end{tabular}
  %}
  \caption{The effect of source languages for comparable corpus creation. These News Crawl corpora are as of 2017 version.
  The number of sentences in each dataset is approximately 20M, respectively. These results are obtained by USMT$_\mathrm{forward}$ in iter 1.}
  \label{multiple_result}
\end{table}

\paragraph{Difference between forward and backward refinements}
%According to Table \ref{result_score}, as the phrase table is updated by back-translation, the correction tends to be passive.
%Table \ref{result_example} shows an example of corrections with increasing iterations.
%USMT$_{\mathrm{backward}}$ suggests many corrections in iter 0, 1 for the source sentence, but most of these corrections are incorrect and change the original meanings.
%However, USMT$_{\mathrm{backward}}$ corrects only one word for each model in iter 2, 3.
To examine how different the refinement methods are, we counted the number of corrections predicted by each method.
The number of USMT$_{\mathrm{forward}}$ in iter 1 and iter 2 is 3,437 and 3,257, respectively,
 whereas that of USMT$_{\mathrm{backward}}$ in iter 1 and iter 2 is 4,092 and 2,789.
As for USMT$_{\mathrm{backward}}$, the number of corrections from iter 1 to iter 2 decreases by 1,303.
\citet{artetxe_usmt} and \citet{lample} reported that the BLEU score \citep{papineni2002} of unsupervised MT with backward-refinement improves with increasing iterations.
In GEC, increasing the iterations of USMT$_{\mathrm{backward}}$ improves the GEC accuracy by predicting less corrections.

\begin{table*}[!t]
  \centering
  \small
  \begin{tabular}{lrrrrrr} \toprule
    Team & TP & FP & FN & P & R & $\mathrm{F_{0.5}}$ \\ \midrule
    UEDIN-MS & 2,312 & 982 & 2,506 & 70.19 & 47.99 & 64.24 \\ 
    Kakao\&Brain & 2,412 & 1,413 & 2,797 & 63.06 & 46.30 & 58.80 \\ 
    LAIX & 1,443 & 884 & 3,175 & 62.01 & 31.25 & 51.81  \\ 
    CAMB-CUED & 1,814 & 1,450 & 2,956 & 55.58 & 38.03 & 50.88  \\
    UFAL, Charles University, Prague & 1,245 & 1,222 & 2,993 & 50.47 & 29.38 & 44.13 \\
    Siteimprove & 1,299 & 1,619 & 3,199 & 44.52 & 28.88 & 40.17 \\
    WebSpellChecker.com & 2,363 & 3,719 & 3,031 & 38.85 & 43.81 & 39.75 \\ 
    TMU & 1,638 & 4,314 & 3,486 & 27.52 & 31.97 & 28.31 \\
    Buffalo & 446 & 1,243 & 3,556 & 26.41 & 11.14 & 20.73 \\ \bottomrule
  \end{tabular}
    \caption{GEC results with W\&I+LOCNESS test data.}
  \label{result_score}
\end{table*}

The GLEU score for USMT$_{\mathrm{forward}}$ is considered higher than that for USMT$_{\mathrm{backward}}$ because the language model makes up for the synthetic target data.
To compare the fluency, the outputs of each best iter on JFLEG were evaluated with the perplexity based on the Common Crawl language model\footnote{http://data.statmt.org/romang/gec-emnlp16/cclm.tgz}.
The perplexity of USMT$_{\mathrm{forward}}$ in iter 1 is 179.23 and that of USMT$_{\mathrm{backward}}$ in iter 1 is 187.49; hence,  the perplexity suggests  USMT$_{\mathrm{forward}}$ produces more likely outputs than USMT$_{\mathrm{backward}}$ under the language model of Common Crawl text.

\paragraph{Effect of the source language}
%We used the translated Finnish News Crawl 2014--2015 as the source-side data.
%However, English News Crawl 2017 was used as the target-side data, so the these dataset are different in taken year.
%Therefore, we examine the effect of the extracted year about source data.
We also examine how source languages of machine translation affect performance.
Table \ref{multiple_result} shows the result in changing the source-side data on CoNLL-14.
%As for the effect of year, the $\mathrm{F_{0.5}}$ score using 2017 Finnish data is 2.00 points higher than that using 2014--2015 data.
%According to this result, it is better to use source and target data which are taken the same year.
The outputs using Finnish data is the best score among various languages;
 the more similar to English the source-side data is, the lower the $\mathrm{F_{0.5}}$ score of the output.   

\begin{table}[!t]
  \centering
  \scalebox{0.9}{
  \small
  \begin{tabular}{lrrrr} \toprule
      & iter & P & R & $\mathrm{F_{0.5}}$    \\ \midrule \midrule
    Unsupervised SMT & 0 & 12.33 & \textbf{16.13} & 12.94 \\
    w/ forward\_refinement & 1 & \textbf{17.59} & 14.63 & \textbf{16.91} \\
     w/o spell check & 2 & 17.30 & 14.15 & 16.56 \\
     & 3 & 16.04 & 14.17 & 15.63 \\ 
     & 4 & 17.06 & 14.01 & 16.35  \\
     & 5 & 15.88 & 13.88 & 15.44 \\ \midrule
     spell check $\to$ SMT & 1 & 20.58 & 18.04 & 20.01 \\
     SMT $\to$ spell check & 1 & 19.42 & 16.86 & 18.85 \\ \bottomrule
  \end{tabular}
  }
    \caption{GEC results with dev data. The bold scores represent the best score without the spell checker.}
  \label{dev_score}
\end{table}

\subsection{W\&I+LOCNESS Results}
Table \ref{result_score} shows the results of the GEC experiments with official test data for W\&I+LOCNESS. 
The $\mathrm{F_{0.5}}$ score for our system (TMU) is 28.31; this score is eighth among the nine teams.
In particular, the number of false positives of our system is 4,314; this is the worst result of all.

%\section{Discussion}
% Effective of refinement
Table \ref{dev_score} shows the results of the dev data listed in Table \ref{data}.
On the dev data, the system of iteration 1 is the best among all.
This tendency is the same as the CoNLL and JFLEG results.
%According to the improvement of iteration from 0 to 1, it is confirmed that the refinement method works well.
%However, it is observed that the system is not improved after iteration 1. 
%The source-side data is fixed, and target-side data is generated from the source-side for each iteration.
%Therefore, the quality of the source-side data is important for this refinement method.
%In this study, we use the automatically translated text as source-side data; thus, it is considered that the quality is not high and the refinement after iteration 1 does not work.

% spell check result
The results of Table \ref{dev_score} confirm that the spell checker works well.
We also investigate the importance of the order; SMT or spell check, which is suitable for the first system for a better result?
As a result, it is better to use the SMT system after using the spell checker.
That is because the source-side data does not include the misspelled words as mentioned above.

\begin{table}[t]
  \centering
  \small
  \begin{tabular}{lrrr} \toprule
      & P & R & $\mathrm{F_{0.5}}$  \\ \midrule \midrule
     Easy2 &  &  &    \\ 
       \hspace{10pt} SPELL & 39.93 & 59.24 & 42.71  \\
        \hspace{10pt} PUNCT & 28.91 & 38.14 & 30.38 \\ \midrule
      Hard2 &    & &  \\
       \hspace{10pt} NOUN & 0.87 & 1.74 & 0.97  \\
       \hspace{10pt} VERB & 2.13 & 0.99 & 1.73 \\ \bottomrule
  \end{tabular}
    \caption{Error types for which our best system corrected errors well or badly on the dev data. Easy2 denotes the easiest two errors, and Hard2 denotes the hardest two errors in terms of the $\mathrm{F_{0.5}}$\footnotemark[11].}
  \label{analysis_error}
\end{table}

\footnotetext[11]{We investigate the frequent error types; the errors occur more than one hundred times in the dev data.}

% error type
Table \ref{analysis_error} shows the error types that our system corrected well or badly on the dev data.
SPELL means the misspell errors; the correction of these errors depends only on the spell checker.
PUNCT means the errors about the punctuation; e.g., `Unfortunately when we...$\to$ Unfortunately, when we...'.
It is considered that our system can correct errors such as these owing to the n-gram co-occurrence knowledge derived from the language models.

In contrast, our system struggled to correct content word errors.
For example, NOUN includes an error like this; `way $\to$ means' and VERB includes an error like this; `watch $\to$ see'.
It is considered that our system is mostly not able to correct the word usage errors based on the context because the phrase table is still noisy.
%the synthetic source data translated by Google Translator is too fluent to include the errors like this.
Although we observed some usage error examples of `watch' in the synthetic source data, our model was not able to replace `watch' to `see' based on the context.
%These word usage errors need to select the proper words from the vocabulary based on the phrase table, but this phrase table of our system did not work well.

\section{Related Work}
\paragraph{Unsupervised Machine Translation}
Studies on unsupervised methods have been conducted for both NMT (\citeauthor{lample}, \citeyear{lample}; \citeauthor{marie-fujita}, \citeyear{marie-fujita}) and SMT \citep{artetxe_usmt}.
%For instance, \citet{artetxe_usmt} proposed an USMT method.
%Other studies such as \citet{lample} and \citet{marie-fujita} show the effectiveness of fine-tuning and initializing of unsupervised NMT (UNMT) with synthetic data generated by USMT.
In this study, we apply the USMT method of \citet{artetxe_usmt} and \citet{marie-fujita} to GEC.
The UNMT method \citep{lample} was ineffective under the GEC setting in our preliminary experiments.

\paragraph{GEC\label{s_gec} with NMT/SMT}
Several studies that introduce sequence-to-sequence models in GEC heavily rely on large amounts of training data.
\citet{ge2018}, who presented state-of-the-art results in GEC, proposed a supervised NMT method trained on corpora of a total 5.4 M sentence pairs.
On the other hand, we mainly use the monolingual corpus and use small learner data as the tuning data.
%because the low resource track does not permit the use of the learner corpora.

Despite the success of NMT, many studies on GEC traditionally use SMT (\citeauthor{susanto2014}, \citeyear{susanto2014}; \citeauthor{junczys2014}, \citeyear{junczys2014}).
These studies apply an off-the-shelf SMT toolkit, Moses, to GEC.
\citet{junczys2014} claimed that the SMT system optimized for BLEU learns to not change the source sentence.
Instead of BLEU, they proposed tuning an SMT system using the $\mathrm{M^2}$ score with annotated development data.
In this study, we also tune the weights with an $\mathrm{F_{0.5}}$ score measured by the $\mathrm{M^2}$ scorer.
% because the official score is an $\mathrm{F_{0.5}}$ score. 
%However, it is difficult to generate adequate development data in the unsupervised setting, as the synthetic development data is too noisy to tune the weights\footnote{Unsupervised tuning with $\mathrm{M^2}$ score was not effective in our preliminary experiments.}.
%The problem of tuning will be addressed in future work.

\paragraph{Low-resource GEC}
\citet{park-levy} proposed a GEC system based on a noisy channel model using an unannotated corpus of learner English.
In contrast, our method does not require an unannotated corpus but requires monolingual corpora. 
%They targeted the spelling, article, preposition and wordform errors.
\citet{lm_gec} built a GEC system using minimally annotated data.
Their model used LM and confusion sets based on the common English error types.
Our method does not require knowledge about the common error types.
\citet{cgmh} proposed a language generation method using Metropolis-Hastings sampling.
This method does not require parallel corpora for training, instead, monolingual data is required.
They evaluate it on a variety of tasks including GEC and report that the GLEU score of 45.5 on JFLEG.
Because we used a parallel corpus for tuning weights, their results cannot be compared with ours.
\citet{copy_gec} reported that a neural GEC model was improved only with denoising auto-encoder, which was trained using a synthetic parallel corpus.
Their parallel corpus was generated by adding artificial errors, such as random deletion of a token, to monolingual data instead of using machine translation. 

%\vspace{-0.5mm}
\section{Conclusion}
%\vspace{-0.5mm}
%We investigated whether the learner corpus needs to be annotated with errors and corrections.
In this paper, we described our GEC system with minimally annotated data.
%for the low resource track of the shared task at BEA2019.
We introduced an unsupervised approach based on SMT for GEC.
This method requires a comparable corpus, so we created a synthetic comparable corpus using Google Translation.
The experimental results demonstrate that our system achieved an $\mathrm{F_{0.5}}$ score of 28.31 points with the W\&I+LOCNESS test data.

%This method achieved an $\mathrm{F_{0.5}}$ score of 22.40 points with the CoNLL-2014 dataset and a GLEU score of 43.85 points with the JFLEG dataset. 
%Its $\mathrm{F_{0.5}}$ score is higher than that of supervised SMT by 2.25 points.
%In addition, we show that updating the phrase table through back-translation makes the system increasingly passive.
%Updating the phrase table using forward refinement results in the generation of fluent outputs.

%Many recent studies on GEC focus on English learners.
%However, there are sentences written by language learners in other languages on language-learning social media platforms, such as Lang-8 and HiNative\footnote{https://hinative.com/}.
%We hope this research paves the way for GEC for low-resourced languages.

%\section*{Acknowledgments}
%This work was partially supported by JSPS Grant-in-Aid for Scientific Research (C) Grant Number JP19K12099.
%We also thank anonymous reviewers for their insightful comments.

\bibliography{GEC}

\begin{thebibliography}{35}
\expandafter\ifx\csname natexlab\endcsname\relax\def\natexlab#1{#1}\fi

\bibitem[{Artetxe et~al.(2018{\natexlab{a}})Artetxe, Labaka, and
  Agirre}]{artetxe_cross}
Mikel Artetxe, Gorka Labaka, and Eneko Agirre. 2018{\natexlab{a}}.
\newblock \href {http://aclweb.org/anthology/P18-1073} {A robust self-learning
  method for fully unsupervised cross-lingual mappings of word embeddings}.
\newblock In \emph{Proc. of ACL}, pages 789--798.

\bibitem[{Artetxe et~al.(2018{\natexlab{b}})Artetxe, Labaka, and
  Agirre}]{artetxe_usmt}
Mikel Artetxe, Gorka Labaka, and Eneko Agirre. 2018{\natexlab{b}}.
\newblock \href {http://aclweb.org/anthology/D18-1399} {Unsupervised
  statistical machine translation}.
\newblock In \emph{Proc. of EMNLP}, pages 3632--3642.

\bibitem[{Bird(2006)}]{nltk}
Steven Bird. 2006.
\newblock \href {http://aclweb.org/anthology/P06-4018} {{NLTK}: The natural
  language toolkit}.
\newblock In \emph{Proc. of COLING/ACL Interactive Presentation Sessions},
  pages 69--72.

\bibitem[{Bojanowski et~al.(2017)Bojanowski, Grave, Joulin, and
  Mikolov}]{fasttext}
Piotr Bojanowski, Edouard Grave, Armand Joulin, and Tomas Mikolov. 2017.
\newblock \href {https://doi.org/10.1162/tacl_a_00051} {Enriching word vectors
  with subword information}.
\newblock \emph{Transactions of the Association for Computational Linguistics},
  5:135--146.

\bibitem[{Bryant and Briscoe(2018)}]{lm_gec}
Christopher Bryant and Ted Briscoe. 2018.
\newblock \href {https://doi.org/10.18653/v1/W18-0529} {Language model based
  grammatical error correction without annotated training data}.
\newblock In \emph{Proc. of BEA}, pages 247--253, New Orleans, Louisiana.

\bibitem[{Bryant et~al.(2019)Bryant, Felice, Andersen, and Briscoe}]{bea2019}
Christopher Bryant, Mariano Felice, {\O}istein~E. Andersen, and Ted Briscoe.
  2019.
\newblock {The {BEA-2019} Shared Task on Grammatical Error Correction}.
\newblock In \emph{Proc. of BEA}.

\bibitem[{Bryant et~al.(2017)Bryant, Felice, and Briscoe}]{errant}
Christopher Bryant, Mariano Felice, and Ted Briscoe. 2017.
\newblock \href {https://doi.org/10.18653/v1/P17-1074} {Automatic annotation
  and evaluation of error types for grammatical error correction}.
\newblock In \emph{Proc. of ACL}, pages 793--805.

\bibitem[{Chelba et~al.(2014)Chelba, Mikolov, Schuster, Ge, Brants, Koehn, and
  Robinson}]{one-billion}
Ciprian Chelba, Tomas Mikolov, Mike Schuster, Qi~Ge, Thorsten Brants, Phillipp
  Koehn, and Tony Robinson. 2014.
\newblock One billion word benchmark for measuring progress in statistical
  language modeling.
\newblock In \emph{Proc. of INTERSPEECH}, pages 2635--2639.

\bibitem[{Dahlmeier and Ng(2012)}]{max_match}
Daniel Dahlmeier and Hwee~Tou Ng. 2012.
\newblock \href {http://aclweb.org/anthology/N12-1067} {Better evaluation for
  grammatical error correction}.
\newblock In \emph{Proc. of NAACL-HLT}, pages 568--572.

\bibitem[{Durrani et~al.(2013)Durrani, Fraser, Schmid, Hoang, and Koehn}]{osm}
Nadir Durrani, Alexander Fraser, Helmut Schmid, Hieu Hoang, and Philipp Koehn.
  2013.
\newblock \href {https://www.aclweb.org/anthology/P13-2071} {Can {Markov}
  models over minimal translation units help phrase-based smt?}
\newblock In \emph{Proc. of ACL}, pages 399--405.

\bibitem[{Dyer et~al.(2013)Dyer, Chahuneau, and Smith}]{dyer2013}
Chris Dyer, Victor Chahuneau, and Noah~A. Smith. 2013.
\newblock \href {http://www.aclweb.org/anthology/N13-1073} {A simple, fast, and
  effective reparameterization of {IBM} model 2}.
\newblock In \emph{Proc. of {NAACL-HLT}}, pages 644--648.

\bibitem[{Ge et~al.(2018)Ge, Wei, and Zhou}]{ge2018}
Tao Ge, Furu Wei, and Ming Zhou. 2018.
\newblock Reaching human-level performance in automatic grammatical error
  correction: An empirical study.
\newblock \emph{arXiv preprint arXiv:1807.01270}.

\bibitem[{Gehring et~al.(2017)Gehring, Auli, Grangier, Yarats, and
  Dauphin}]{convs2s}
Jonas Gehring, Michael Auli, David Grangier, Denis Yarats, and Yann~N Dauphin.
  2017.
\newblock Convolutional sequence to sequence learning.
\newblock In \emph{Proc. of ICML}, pages 1243--1252.

\bibitem[{Granger(1998)}]{locness}
Sylviane Granger. 1998.
\newblock {The computer learner corpus: A versatile new source of data for SLA
  research.}
\newblock In Sylviane Granger, editor, \emph{{Learner English on Computer}},
  pages 3--18. Addison Wesley Longman, London and New York.

\bibitem[{Grundkiewicz and Junczys-Dowmunt(2018)}]{hybrid-gec}
Roman Grundkiewicz and Marcin Junczys-Dowmunt. 2018.
\newblock \href {https://doi.org/10.18653/v1/N18-2046} {Near human-level
  performance in grammatical error correction with hybrid machine translation}.
\newblock In \emph{Proc. of NAACL-HLT}, pages 284--290.

\bibitem[{Heafield(2011)}]{kenlm}
Kenneth Heafield. 2011.
\newblock \href {http://aclweb.org/anthology/W11-2123} {{KenLM}: Faster and
  smaller language model queries}.
\newblock In \emph{Proc. of WMT}, pages 187--197.

\bibitem[{Junczys-Dowmunt and Grundkiewicz(2014)}]{junczys2014}
Marcin Junczys-Dowmunt and Roman Grundkiewicz. 2014.
\newblock The {AMU} system in the {CoNLL-2014} shared task: Grammatical error
  correction by data-intensive and feature-rich statistical machine
  translation.
\newblock In \emph{Proc. of CoNLL}, pages 25--33.

\bibitem[{Koehn et~al.(2007)Koehn, Hoang, Birch, Callison-Burch, Federico,
  Bertoldi, Cowan, Shen, Moran, Zens, Dyer, Bojar, Constantin, and
  Herbst}]{moses}
Philipp Koehn, Hieu Hoang, Alexandra Birch, Chris Callison-Burch, Marcello
  Federico, Nicola Bertoldi, Brooke Cowan, Wade Shen, Christine Moran, Richard
  Zens, Chris Dyer, Ondrej Bojar, Alexandra Constantin, and Evan Herbst. 2007.
\newblock \href {http://aclweb.org/anthology/P07-2045} {Moses: Open source
  toolkit for statistical machine translation}.
\newblock In \emph{Proc. of ACL Demo Sessions}, pages 177--180.

\bibitem[{Lample et~al.(2018)Lample, Ott, Conneau, Denoyer, and
  Ranzato}]{lample}
Guillaume Lample, Myle Ott, Alexis Conneau, Ludovic Denoyer, and Marc'Aurelio
  Ranzato. 2018.
\newblock \href {http://aclweb.org/anthology/D18-1549} {Phrase-based {\&}
  neural unsupervised machine translation}.
\newblock In \emph{Proc. of EMNLP}, pages 5039--5049.

\bibitem[{Marie and Fujita(2018)}]{marie-fujita}
Benjamin Marie and Atsushi Fujita. 2018.
\newblock \href {http://arxiv.org/abs/1810.12703} {Unsupervised neural machine
  translation initialized by unsupervised statistical machine translation}.
\newblock \emph{arXiv preprint arXiv:1810.12703}.

\bibitem[{Miao et~al.(2019)Miao, Zhou, Mou, Yan, and Li}]{cgmh}
Ning Miao, Hao Zhou, Lili Mou, Rui Yan, and Lei Li. 2019.
\newblock {CGMH}: Constrained sentence generation by metropolis-hastings
  sampling.
\newblock In \emph{Proc. of AAAI}.

\bibitem[{Mikolov et~al.(2013)Mikolov, Chen, Corrado, and
  Dean}]{mikolov2013_skip}
Tomas Mikolov, Kai Chen, Greg Corrado, and Jeffrey Dean. 2013.
\newblock Efficient estimation of word representations in vector space.
\newblock In \emph{ICLR Workshop}.

\bibitem[{Mizumoto et~al.(2011)Mizumoto, Komachi, Nagata, and
  Matsumoto}]{mizumoto2011}
Tomoya Mizumoto, Mamoru Komachi, Masaaki Nagata, and Yuji Matsumoto. 2011.
\newblock Mining revision log of language learning {SNS} for automated
  {Japanese} error correction of second language learners.
\newblock In \emph{Proc. of IJCNLP}, pages 147--155.

\bibitem[{Napoles et~al.(2015)Napoles, Sakaguchi, Post, and Tetreault}]{gleu}
Courtney Napoles, Keisuke Sakaguchi, Matt Post, and Joel Tetreault. 2015.
\newblock \href {https://doi.org/10.3115/v1/P15-2097} {Ground truth for
  grammatical error correction metrics}.
\newblock In \emph{Proc. of ACL-IJCNLP}, pages 588--593.

\bibitem[{Napoles et~al.(2017)Napoles, Sakaguchi, and Tetreault}]{jfleg}
Courtney Napoles, Keisuke Sakaguchi, and Joel Tetreault. 2017.
\newblock \href {http://aclweb.org/anthology/E17-2037} {{JFLEG}: A fluency
  corpus and benchmark for grammatical error correction}.
\newblock In \emph{Proc. of EACL}, pages 229--234.

\bibitem[{Ng et~al.(2014)Ng, Wu, Briscoe, Hadiwinoto, Susanto, and
  Bryant}]{ng2014}
Hwee~Tou Ng, Siew~Mei Wu, Ted Briscoe, Christian Hadiwinoto, Raymond~Hendy
  Susanto, and Christopher Bryant. 2014.
\newblock The {CoNLL}-2014 shared task on grammatical error correction.
\newblock In \emph{Proc. of {CoNLL} Shared Task}, pages 1--14.

\bibitem[{Ng et~al.(2013)Ng, Wu, Wu, Hadiwinoto, and Tetreault}]{conll13}
Hwee~Tou Ng, Siew~Mei Wu, Yuanbin Wu, Christian Hadiwinoto, and Joel Tetreault.
  2013.
\newblock \href {http://aclweb.org/anthology/W13-3601} {The {CoNLL-2013} shared
  task on grammatical error correction}.
\newblock In \emph{Proc. of CoNLL}, pages 1--12.

\bibitem[{Och(2003)}]{mert}
Franz~Josef Och. 2003.
\newblock \href {http://aclweb.org/anthology/P03-1021} {Minimum error rate
  training in statistical machine translation}.
\newblock In \emph{Proc. of ACL}, pages 160--167.

\bibitem[{Papineni et~al.(2002)Papineni, Roukos, Ward, and Zhu}]{papineni2002}
Kishore Papineni, Salim Roukos, Todd Ward, and Wei-Jing Zhu. 2002.
\newblock \href {http://www.aclweb.org/anthology/P02-1040} {{BLEU}: a method
  for automatic evaluation of machine translation}.
\newblock In \emph{Proc. of {ACL}}, pages 311--318.

\bibitem[{Park and Levy(2011)}]{park-levy}
Y.~Albert Park and Roger Levy. 2011.
\newblock \href {https://www.aclweb.org/anthology/P11-1094} {Automated whole
  sentence grammar correction using a noisy channel model}.
\newblock In \emph{Proc. of ACL}, pages 934--944, Portland, Oregon, USA.

\bibitem[{Sennrich et~al.(2016{\natexlab{a}})Sennrich, Haddow, and
  Birch}]{back-translation}
Rico Sennrich, Barry Haddow, and Alexandra Birch. 2016{\natexlab{a}}.
\newblock \href {https://doi.org/10.18653/v1/P16-1009} {Improving neural
  machine translation models with monolingual data}.
\newblock In \emph{Proc. of ACL}, pages 86--96.

\bibitem[{Sennrich et~al.(2016{\natexlab{b}})Sennrich, Haddow, and Birch}]{bpe}
Rico Sennrich, Barry Haddow, and Alexandra Birch. 2016{\natexlab{b}}.
\newblock \href {https://doi.org/10.18653/v1/P16-1162} {Neural machine
  translation of rare words with subword units}.
\newblock In \emph{Proc. of ACL}, pages 1715--1725.

\bibitem[{Susanto et~al.(2014)Susanto, Phandi, and Ng}]{susanto2014}
Raymond~Hendy Susanto, Peter Phandi, and Hwee~Tou Ng. 2014.
\newblock \href {https://doi.org/10.3115/v1/D14-1102} {System combination for
  grammatical error correction}.
\newblock In \emph{Proc. of EMNLP}, pages 951--962.

\bibitem[{Xie et~al.(2018)Xie, Genthial, Xie, Ng, and Jurafsky}]{xie2018}
Ziang Xie, Guillaume Genthial, Stanley Xie, Andrew Ng, and Dan Jurafsky. 2018.
\newblock \href {https://doi.org/10.18653/v1/N18-1057} {Noising and denoising
  natural language: Diverse backtranslation for grammar correction}.
\newblock In \emph{Proc. of NAACL-HLT}, pages 619--628.

\bibitem[{Zhao et~al.(2019)Zhao, Wang, Shen, Jia, and Liu}]{copy_gec}
Wei Zhao, Liang Wang, Kewei Shen, Ruoyu Jia, and Jingming Liu. 2019.
\newblock \href {https://www.aclweb.org/anthology/N19-1014} {Improving
  grammatical error correction via pre-training a copy-augmented architecture
  with unlabeled data}.
\newblock In \emph{Proc. of NAACL-HLT}, pages 156--165.

\end{thebibliography}
\bibliographystyle{acl_natbib}

\end{document}